\documentclass[11pt,a4paper]{article}
\usepackage[hyperref]{acl2018}
\usepackage{times}
\usepackage{latexsym}
\usepackage{amssymb}
\usepackage{stackengine}
\usepackage{multirow}
\usepackage[fleqn]{amsmath}
\usepackage{graphicx}
\usepackage{hhline}
\usepackage{array}
\usepackage{booktabs}
\graphicspath{{figures/}}
\renewcommand{\vec}[1]{\boldsymbol{#1}} 

\newcommand{\rowgroup}[1]{\hspace{-1em}#1}
\usepackage{amsfonts}
\newcommand\blfootnote[1]{%
 \begingroup
 \renewcommand\thefootnote{}\footnote{#1}%
 \addtocounter{footnote}{-1}%
 \endgroup
}

\usepackage{url}

\aclfinalcopy 

\newcommand\vtheta{\pmb{\theta}}
\newcommand\vphi{\pmb{\phi}}

\title{Exploring Textual and Speech information in Dialogue Act Classification with Speaker Domain Adaptation}
\author{
\begin{tabular}{ccc}
\bf{Xuanli He}$^*$ &  \bf{Quan Hung Tran}$^*$ & \bf{William Havard}\tabularnewline
Monash University & Monash University & Univ. Grenoble Alpes\tabularnewline
xuanli.he1@monash.edu & hung.tran@monash.edu&william.havard@gmail.com
\\\\
\textbf{Laurent Besacier} & \textbf{Ingrid Zukerman} & \textbf{Gholamreza Haffari}\tabularnewline
Univ. Grenoble Alpes & Monash University & Monash University\tabularnewline
laurent.besacier@imag.fr &ingrid.zukerman@monash.edu&gholamreza.haffari@monash.edu
\end{tabular}}

\date{}

\begin{document}
 \maketitle
 \blfootnote{$^*$equal contribution}
\begin{abstract}
In spite of the recent success of Dialogue Act (DA) classification, 
the majority of prior works focus on text-based classification with oracle transcriptions, i.e. human transcriptions, 
instead of Automatic Speech Recognition (ASR)'s transcriptions. 
In spoken dialog systems, however, the agent would only have access to noisy ASR transcriptions, which may further suffer  performance degradation due to domain shift. 
In this paper, we explore the effectiveness of using both acoustic and textual signals, either oracle or ASR transcriptions, 
and investigate speaker domain adaptation for DA classification.
Our multimodal model proves to be superior to the unimodal models, particularly when the oracle transcriptions are not available. 
We also propose an effective  method for speaker domain adaptation,  which
achieves competitive results. 
  
\end{abstract}

\section{Introduction}
Dialogue Act (DA) classification is a sequence-labelling task, mapping a sequence
of utterances to their corresponding DAs. Since DA classification plays an important
role in understanding spontaneous dialogue \cite{DBLP:journals/corr/cs-CL-0006023},
numerous techniques have been proposed to capture the semantic correlation between
utterances and DAs.

Earlier on, statistical techniques  such as Hidden Markov Models (HMMs) were widely used to recognise DAs \cite{DBLP:journals/corr/cs-CL-0006023,doi:10.1142/S1469026810002926}.
Recently, due to the enormous success of neural networks in sequence labeling/transduction tasks
\cite{DBLP:journals/corr/SutskeverVL14,DBLP:journals/corr/BahdanauCB14,
DBLP:journals/corr/NallapatiXZ16,Popov2016}, several recurrent neural network (RNN) based architectures 
  have been proposed
to conduct DA classification, resulting in  promising outcomes \cite{DBLP:journals/corr/JiHE16,
DBLP:journals/corr/ShenL16,tran2017hierarchical}.

Despite the success of previous work in DA classification, there are still several fundamental
issues. Firstly, most of the previous works rely on transcriptions \cite{DBLP:journals/corr/JiHE16,
DBLP:journals/corr/ShenL16,tran2017hierarchical}. Fewer of these focus on combining speech and textual signals \cite{doi:10.1142/S1469026810002926}, and even then,
the textual signals in these works come from
the oracle transcriptions. 
We argue that in the context
of a spoken dialog system, oracle transcriptions of
utterances are usually not available, i.e. the agent does not have access to the human transcriptions to produce answers. 
Speech  and textual data
complement each other, especially when textual data is from ASR systems rather than oracle transcripts. 
Furthermore, domain adaptation in text or speech-based DA classification is relatively under-investigated. As shown in our experiments, DA classification models perform much worse when they are applied to new speakers. 

In this paper, we explore the effectiveness of using both acoustic and textual signals, 
and investigate speaker domain adaptation for DA classification.
We present a multimodal model to combine text and speech signals, which proves to be superior to the unimodal models, 
particularly when the oracle transcriptions are not available. 
Our experiments are conducted on Switchboard \cite{godfrey1992switchboard,jurafsky97switchboard} and Maptask \cite{anderson1991hcrc}.
Moreover, we propose an effective  method for speaker domain adaptation,  which learns a suitable encoder for the new domain giving rise to representations similar to those in the source domain.  
Our domain adaptation does not need any labeled data from the target domain, and significantly outperforms the unadapted model. 
%
%

\begin{figure*}[t]
\includegraphics[scale=0.23]{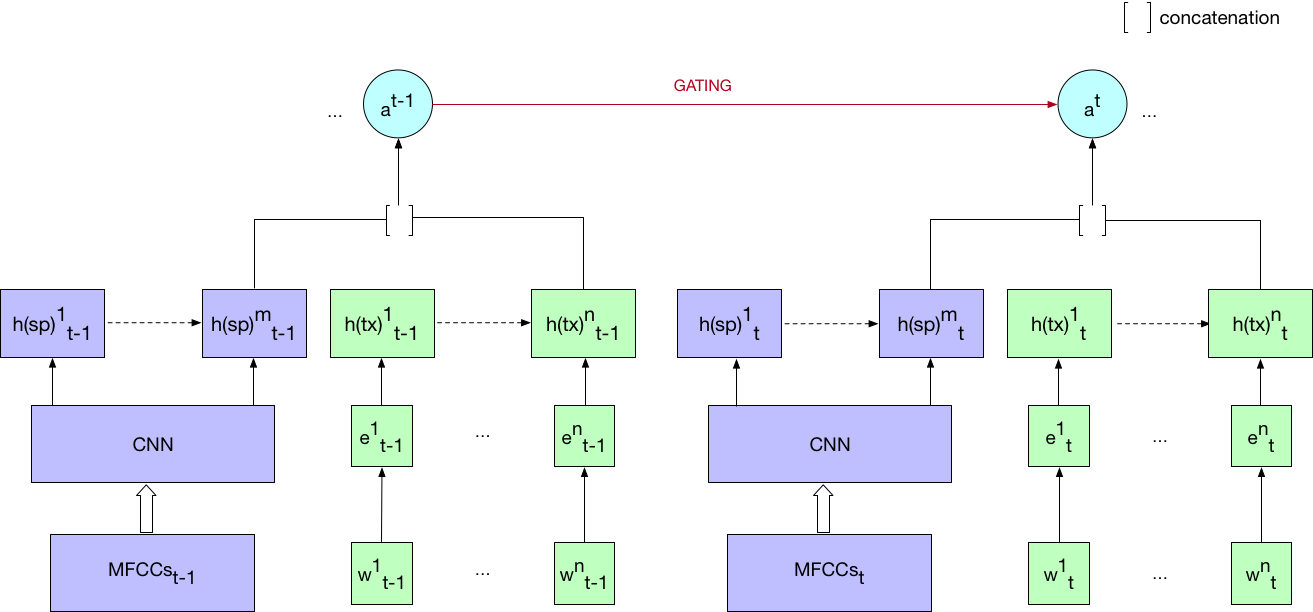}
\centering
\caption{The multimodal model. For the utterance  $t$, the left and right sides are encoding speech and text, respectively. }
\label{fig:1}
\end{figure*}

\section{Model Description} \label{sec:2}

In this section, we describe the basic structure of our model, which combines the textual
and speech modalities. We also introduce a representation learning approach using adversarial ideas to 
tackle the domain adaptation problem.

\subsection{Our Multimodal Model} \label{sec:21}

A conversation is comprised of a sequence of utterances $\vec{u}_1,...,\vec{u}_T$, and each utterance $\vec{u}_t$
is labeled with a DA $a_t$. An utterance could include text, speech or both.
We focus on online DA classification, and our classification model attempts to directly model the conditional
probability $p(\vec{a}_{1:T}|\vec{u}_{1:T})$ decomposed as follows:

\begin{align} \label{eq:1}
p(\vec{a}_{1:T}|\vec{u}_{1:T})&=\prod_{t=1}^T{p(a_t|a_{t-1},\vec{u}_t)}.
\end{align}
To incorporate the previous DA information and tackle the label-bias problem, we adopt the uncertainty propagation
architecture \cite{tran2017preserving}. The conditional probability term in Eqn. \ref{eq:1} is computed as follows: 
{\setlength{\mathindent}{0.5cm}
\begin{align*} \label{eq:2}
 \begin{split}
& a_t|a_{t-1},\vec{u}_t \sim \vec{q}_t \\
& \vec{q}_t =  softmax(\overline{\vec{W}}\cdot \vec{c}(\vec{u}_t) + \overline{\bf b}) \\
& \overline{\vec{W}} = \sum_a \vec{q}_{t-1}(a) \vec{W}^a \ \ , \ \  
 \overline{\vec{b}} = \sum_a \vec{q}_{t-1}(a) \vec{b}^a 
%
  \end{split}
\end{align*}
where $\vec{W}^{a}$ and $\vec{b}^{a}$ are DA-specific parameters gated on the  DA $a$, $\vec{c}(\vec{u}_t)$
is the encoding of the utterance $\vec{u}_t$, and  $\vec{q}_{t-1}$ represents the uncertainty distribution over the DAs at the time step $t-1$.

\smallskip
\noindent
\textbf{Text Utterance.}  An utterance $\vec{u}_t$ includes a list of words $w_t^1,...,w_t^n$. The word $w_t^i$
is embedded by $\vec{x}_t^i=\vec{e}(w_t^i)$ where $\vec{e}$ is the embedding table. 

\smallskip
\noindent
\textbf{Speech Utterance.} We apply a frequency-based transformation on raw speech
signals to acquire Mel-frequency cepstral coefficients (MFCCs), which have been very effective in speech recognition \cite{mohamed2012understanding}.
To learn the context-specific  features of the speech signal, a convolutional neural network (CNN) is employed
over MFCCs:
{\setlength{\mathindent}{1cm}
\begin{align*}
\vec{x'}_t^1,...,\vec{x'}_t^m = CNN(\vec{s}_t^1,...,\vec{s}_t^k)
\end{align*}
where $\vec{s}_t^i$ is a MFCC feature vector at the position $i$ for the $t$-th utterance.

\smallskip
\noindent
\textbf{Encoding of Text+Speech.} We employ two RNNs with LSTM units to encode the text and speech sequences 
 an utterance $\vec{u}_t$:
 \begin{align*}
\vec{c}(\vec{u}_t)^{tx} = RNN_{\vtheta}(\vec{x}_t^1,...,\vec{x}_t^n)\\
\vec{c}(\vec{u}_t)^{sp} = RNN_{\vtheta'}(\vec{x'}_t^1,...,\vec{x'}_t^m).
\end{align*}
where  the encoding of the text $\vec{c}(\vec{u}_t)^{tx}$
and  speech $\vec{c}(\vec{u}_t)^{sp}$ are the last hidden states of the corresponding RNNs whose parameters are denoted by  $\vtheta$ and $\vtheta'$. 
The distributed representation $\vec{c}(\vec{u}_t)$ of the utterance $\vec{u}_t$ is then the concatenation
of $\vec{c}(\vec{u}_t)^{tx}$ and $\vec{c}(\vec{u}_t)^{sp}$.

\subsection{Speaker Domain Adaptation } \label{sec:22}
Different people tend to speak differently. This  creates a problem for DA classification systems, as unfamiliar speech signals might not be
recognised properly. 
In our preliminary experiments, the performance of DA classification on speakers that appear in the training
set is found to be significantly higher than that of unknown speakers in the test set. This motivates  to explore the problem of speaker domain adaptation in DA classification.

\begin{figure*}[t]
\includegraphics[scale=0.24]{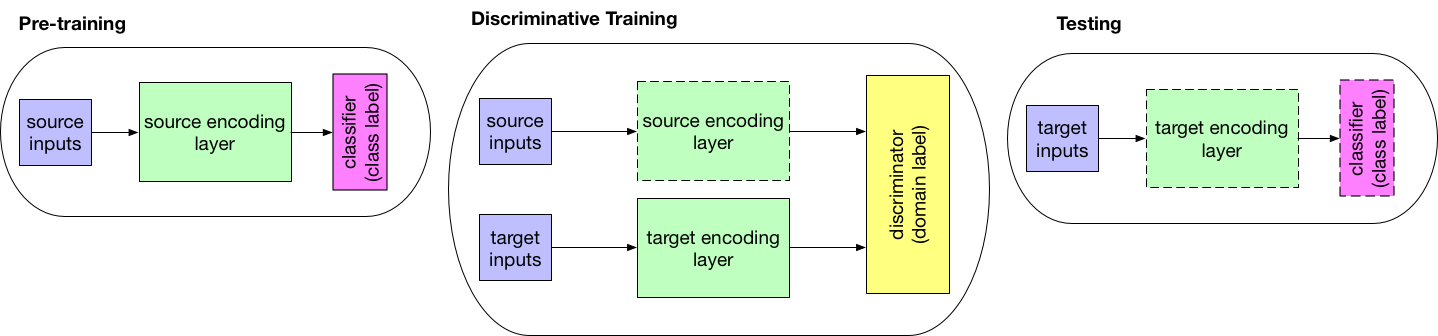}
\centering
\caption{Overview of discriminative model. Dashed lines indicate frozen parts}
\label{fig:2}
\end{figure*}

We assume we have a large amount of labelled source data pair $\{X_{src},Y_{src}\}$, and a small amount of unlabelled target data $X_{trg}$, where an utterance $\vec{u} \in X$ includes both speech and text parts.
Inspired by \citet{DBLP:journals/corr/TzengHSD17}, our goal is to learn a target domain encoder  which can fool a domain classifier $C_{\phi}$ in distinguishing whether the utterance belongs to the source or target domain. 
Once the target encoder is trained to produce representations which look like those coming from the source domain, 
the target encoder can be used together with other components of the source DA prediction model to predict DAs for the target domain (see Figure 2). 

We use a 1-layer feed-forward network as the domain classifier:
$$C_{\vphi}(\vec{r}) = \sigma(\vec{W}_C \cdot \vec{r} + {b}_C)$$
where the classifier produces the probability of the input representation $\vec{r}$ belonging to the source domain, and 
$\vphi$ denotes the classifier parameters $\{\vec{W}_C, {b}_C\}$. 
Let the target and source domain encoders are denoted by $\vec{c}_{trg}(\vec{u}_{trg})$ and $\vec{c}_{src}(\vec{u}_{trg})$, respectively.  
The training objective of the domain classifier is: 
{\setlength{\mathindent}{0cm}
\begin{align*}
&\mathop{min}_{\vphi}\quad\mathcal{L}_1(X_{src},X_{trg}, C_{\vphi})=\\
                                                                    &\qquad\qquad-\mathbb{E}_{\vec{u}\sim X_{src}}[\log C_{\vphi}(\vec{c}_{src}(\vec{u}))]\\
                                                                      &\qquad\qquad-\mathbb{E}_{\vec{u}\sim X_{trg}}[1-\log C_{\vphi}(\vec{c}_{trg}(\vec{u}))].                                                   
\end{align*}
As mentioned before, we keep the source encoder fixed and train the parameters of the target domain encoder. 
The training objective of the target domain encoder is
{\setlength{\mathindent}{0cm}
\begin{align*}
&\mathop{min}_{\vtheta'_{trg}}\quad\mathcal{L}_2(X_{trg}, C_{\vphi})=\\
                                                                      &\qquad\qquad-\mathbb{E}_{\vec{u} \sim X_{trg}}[\log C_{\vphi}(\vec{c}_{trg}(\vec{u}))]                                             
\end{align*}
where the optimisation is performed over the speech RNN parameters $\vtheta'_{trg}$ of the target encoder. 
We also tried to optimise other parameters (i.e. CNN parameters,
word embeddings and text RNN parameters), but the performance is similar to the speech RNN only. This is possibly because the major
difference between source  and target domain data is due to the speech signals. We alternate between optimising $\mathcal{L}_1$ and $\mathcal{L}_2$ using Adam \cite{DBLP:journals/corr/KingmaB14}  until a training condition is met. 

\section{Experiments} \label{sec:3}
\subsection{Datasets}
We test our models on two datasets: the MapTask Dialog Act data \cite{anderson1991hcrc} and
the Switchboard Dialogue Act data \cite{jurafsky97switchboard}.

\smallskip
\noindent
\textbf{MapTask dataset} This dataset consist of 128 conversations labelled with 13 DAs. We randomly partition this data into 80\%
training, 10\% development and 10\% test sets, having 103, 12 and 13 conversations respectively.

\smallskip
\noindent
\textbf{Switchboard dataset} There are 1155 transcriptions of telephone conversations in this dataset, and each utterance falls into one of 
 42 DAs. We follow the setup proposed by \citet{DBLP:journals/corr/cs-CL-0006023}: 1115 conversations for training,
 21 for development and 19 for testing. Since we do not have access to the original recordings of Switchboard dataset, we use synthetic
 speeches generated by a text-to-speech (TTS) system.

\subsection{Results}
\paragraph{In-Domain Evaluation.} Unlike most prior work \cite{DBLP:journals/corr/JiHE16,DBLP:journals/corr/ShenL16,tran2017hierarchical}, we use ASR transcripts, 
produced by the CMUSphinx ASR system, rather than the oracle text. 
We argue that most dialogues in the real world are in the speech format,
 thus our setup is closer to the real-life  scenario.

As shown in Tables \ref{tab:1} and \ref{tab:2}, our multimodal model outperforms strong baselines
on Switchboard and MapTask datasets, when using the ASR transcriptions. 
When using the oracle text, the information from the speech signal does not lead to further improvement though, possibly 
due to the existence of acoustic features (such as tones, question markers etc) in the high quality transcriptions. 
 On MapTask, there is a large gap between oracle-based and ASR-based models. 
 This degradation is mainly caused by the poor quality  acoustic signals in MapTask, making ASR ineffective compared to directly predicting DAs from the speech signal.

\begin{table}[ht]
\begin{center}
\begin{tabular}{>{\quad}lc}
\toprule
Models & Accuracy\\
\midrule
\rowgroup{Oracle text}\\
\citet{DBLP:journals/corr/cs-CL-0006023} & 71.00\%\\
\citet{DBLP:journals/corr/ShenL16}&72.60\%\\
\citet{tran2017hierarchical} &74.50\%\\
Text only (ours) & 74.97\%\\
Text+Speech (ours) & 74.98\%\\
\midrule
\rowgroup{Speech and ASR}\\
Speech only & 59.71\%\\
Text only (ASR) & 66.39\%\\
Text+Speech (ASR)&\textbf{68.25}\%\\
\bottomrule
\end{tabular}
\end{center}
\caption{Results of different models on Switchboard data.}
\label{tab:1}
\end{table}

\begin{table}[ht]
\begin{center}
\begin{tabular}{>{\quad}lc}
\toprule
Models & Accuracy\\
\midrule
\rowgroup{Oracle text}\\
\citet{doi:10.1142/S1469026810002926}&55.40\%\\
\citet{tran2017hierarchical} &61.60\%\\
Text only (ours) & \textbf{61.73}\%\\
Text+Speech (ours) & 61.67\%\\
\midrule
\rowgroup{Speech and ASR}\\
Speech only & 39.32\%\\
Text only (ASR) & 38.10\%\\
Text+Speech (ASR)&\textbf{39.39}\%\\
\bottomrule
\end{tabular}
\end{center}
\caption{Results of different models on MapTask data.}
\label{tab:2}
\end{table}

\paragraph{Out-of-Domain Evaluation.} 
We  evaluate our domain adaptation model on the out of domain data on Switchboard. 
Our training data comprises of five known speakers, whereas development and test sets include data from   three new speakers. 
The speeches for these 8 speakers are generated by a TTS system from the oracle transcriptions.
 
As described in Section \ref{sec:22}, we pre-train our speech models on the labeled
training data from the 5 known speakers, then train speech encoders for the new speakers using speeches from both known and new speakers. 
During domain adaptation, the five known
speakers are marked as the source domain,  while the three new speakers are treated as the target domains. 
For  domain adaptation with unlabelled data, the DA tags of both the source  and target domains are removed. 
We test the source-only model and the domain adaptation models merely on the three new speakers in test data.
As shown in Table \ref{tab:3}, compared with the source-only model, the domain adaptation strategy improves the
performance of speech-only and text+speech models, consistently and substantially.

\begin{table}[ht]
\begin{center}
\begin{tabular}{rcc}
 \hline    
    Methods &  Speech & Text+Speech \\
    \hline
        Unadapted  &   48.73\%  &63.57\%\\
        Domain Adapted &          54.37\% &  67.21\%\\
        \hline 
                Supervised Learning &  56.19 \%  & 68.04\%\\
    \hline
\end{tabular}
\end{center}
\caption{Experimental results of the unadapted (i.e. source-only) and domain adapted models using unlabeled data on Switchboard, as well as  the supervised learning upperbound. }
\label{tab:3}
\end{table}

To assess the effectiveness of our domain adaptation architecture, we compare it with  the supervised learning scenario where the model has access to labeled data from all speakers during training. 
To do this, we randomly add two thirds of labelled development data of new speakers to the training set,
and apply the trained model to the test set. 
%
The supervised learning scenario is an upper-bound to our domain adaptation approach, as it makes use of labeled data; see the results in the last row of Table \ref{tab:3}. 
However, the gap between supervised learning and  domain adaptation is not big compared to that between the adapted and unadapted models, showing that our domain adaption technique has been effective.

\section{Conclusion}
In this paper, we have proposed a multimodal model to combine textual and acoustic signals for DA prediction. 
We have demonstrated that the our model 
exceeds unimodal models, especially when oracle transcriptions do not exist. 
In addition, we have proposed  an effective domain adaptation technique in order to adapt our multimodal DA prediction model to new speakers.
%
\bibliography{acl2018}

\begin{thebibliography}{}
\expandafter\ifx\csname natexlab\endcsname\relax\def\natexlab#1{#1}\fi

\bibitem[{Anderson et~al.(1991)Anderson, Bader, Bard, Boyle, Doherty, Garrod,
  Isard, Kowtko, McAllister, Miller et~al.}]{anderson1991hcrc}
Anne~H Anderson, Miles Bader, Ellen~Gurman Bard, Elizabeth Boyle, Gwyneth
  Doherty, Simon Garrod, Stephen Isard, Jacqueline Kowtko, Jan McAllister, Jim
  Miller, et~al. 1991.
\newblock The hcrc map task corpus.
\newblock {\em Language and speech\/} 34(4):351--366.

\bibitem[{Bahdanau et~al.(2014)Bahdanau, Cho, and
  Bengio}]{DBLP:journals/corr/BahdanauCB14}
Dzmitry Bahdanau, Kyunghyun Cho, and Yoshua Bengio. 2014.
\newblock Neural machine translation by jointly learning to align and
  translate.
\newblock {\em CoRR\/} abs/1409.0473.

\bibitem[{Godfrey et~al.(1992)Godfrey, Holliman, and
  McDaniel}]{godfrey1992switchboard}
John~J Godfrey, Edward~C Holliman, and Jane McDaniel. 1992.
\newblock Switchboard: Telephone speech corpus for research and development.
\newblock In {\em Acoustics, Speech, and Signal Processing, 1992. ICASSP-92.,
  1992 IEEE International Conference on\/}. IEEE, volume~1, pages 517--520.

\bibitem[{Ji et~al.(2016)Ji, Haffari, and
  Eisenstein}]{DBLP:journals/corr/JiHE16}
Yangfeng Ji, Gholamreza Haffari, and Jacob Eisenstein. 2016.
\newblock A latent variable recurrent neural network for discourse relation
  language models.
\newblock {\em CoRR\/} abs/1603.01913.

\bibitem[{Julia et~al.(2010)Julia, Iftekharuddin, and
  Islam}]{doi:10.1142/S1469026810002926}
Fatema~N. Julia, Khan~M. Iftekharuddin, and Atiq~U. Islam. 2010.
\newblock Dialog act classification using acoustic and discourse information of
  maptask data.
\newblock {\em International Journal of Computational Intelligence and
  Applications\/} 09(04):289--311.

\bibitem[{Jurafsky et~al.(1997)Jurafsky, Shriberg, and
  Biasca}]{jurafsky97switchboard}
D.~Jurafsky, E.~Shriberg, and D.~Biasca. 1997.
\newblock {Switchboard SWBD-DAMSL shallow-discourse-function annotation coders
  manual}.
\newblock Technical Report Draft 13, University of Colorado, Institute of
  Cognitive Science.

\bibitem[{Kingma and Ba(2014)}]{DBLP:journals/corr/KingmaB14}
Diederik~P. Kingma and Jimmy Ba. 2014.
\newblock \href{http://arxiv.org/abs/1412.6980}{Adam: {A} method for stochastic
  optimization}.
\newblock {\em CoRR\/} abs/1412.6980.
\newblock
  \href{http://arxiv.org/abs/1412.6980}{http://arxiv.org/abs/1412.6980}.

\bibitem[{Mohamed et~al.(2012)Mohamed, Hinton, and
  Penn}]{mohamed2012understanding}
Abdel-rahman Mohamed, Geoffrey Hinton, and Gerald Penn. 2012.
\newblock Understanding how deep belief networks perform acoustic modelling.
\newblock In {\em Acoustics, Speech and Signal Processing (ICASSP), 2012 IEEE
  International Conference on\/}. IEEE, pages 4273--4276.

\bibitem[{Nallapati et~al.(2016)Nallapati, Xiang, and
  Zhou}]{DBLP:journals/corr/NallapatiXZ16}
Ramesh Nallapati, Bing Xiang, and Bowen Zhou. 2016.
\newblock Sequence-to-sequence rnns for text summarization.
\newblock {\em CoRR\/} abs/1602.06023.

\bibitem[{Popov(2016)}]{Popov2016}
Alexander Popov. 2016.
\newblock {\em Deep Learning Architecture for Part-of-Speech Tagging with Word
  and Suffix Embeddings\/}, Springer International Publishing, Cham, pages
  68--77.

\bibitem[{Shen and Lee(2016)}]{DBLP:journals/corr/ShenL16}
Sheng{-}syun Shen and Hung{-}yi Lee. 2016.
\newblock Neural attention models for sequence classification: Analysis and
  application to key term extraction and dialogue act detection.
\newblock {\em CoRR\/} abs/1604.00077.

\bibitem[{Stolcke et~al.(2000)Stolcke, Ries, Coccaro, Shriberg, Bates,
  Jurafsky, Taylor, Martin, Ess{-}Dykema, and
  Meteer}]{DBLP:journals/corr/cs-CL-0006023}
Andreas Stolcke, Klaus Ries, Noah Coccaro, Elizabeth Shriberg, Rebecca~A.
  Bates, Daniel Jurafsky, Paul Taylor, Rachel Martin, Carol~Van Ess{-}Dykema,
  and Marie Meteer. 2000.
\newblock Dialogue act modeling for automatic tagging and recognition of
  conversational speech.
\newblock {\em CoRR\/} cs.CL/0006023.

\bibitem[{Sutskever et~al.(2014)Sutskever, Vinyals, and
  Le}]{DBLP:journals/corr/SutskeverVL14}
Ilya Sutskever, Oriol Vinyals, and Quoc~V. Le. 2014.
\newblock Sequence to sequence learning with neural networks.
\newblock {\em CoRR\/} abs/1409.3215.

\bibitem[{Tran et~al.(2017{\natexlab{a}})Tran, Zukerman, and
  Haffari}]{tran2017hierarchical}
Quan~Hung Tran, Ingrid Zukerman, and Gholamreza Haffari. 2017{\natexlab{a}}.
\newblock A hierarchical neural model for learning sequences of dialogue acts.
\newblock In {\em Proceedings of the 15th Conference of the European Chapter of
  the Association for Computational Linguistics: Volume 1, Long Papers\/}.
  volume~1, pages 428--437.

\bibitem[{Tran et~al.(2017{\natexlab{b}})Tran, Zukerman, and
  Haffari}]{tran2017preserving}
Quan~Hung Tran, Ingrid Zukerman, and Gholamreza Haffari. 2017{\natexlab{b}}.
\newblock Preserving distributional information in dialogue act classification.
\newblock In {\em Proceedings of the 2017 Conference on Empirical Methods in
  Natural Language Processing\/}. pages 2141--2146.

\bibitem[{Tzeng et~al.(2017)Tzeng, Hoffman, Saenko, and
  Darrell}]{DBLP:journals/corr/TzengHSD17}
Eric Tzeng, Judy Hoffman, Kate Saenko, and Trevor Darrell. 2017.
\newblock Adversarial discriminative domain adaptation.
\newblock {\em CoRR\/} abs/1702.05464.

\end{thebibliography}
\bibliographystyle{acl_natbib}

\
\end{document}